\documentclass[letterpaper, 10 pt, conference]{ieeeconf}  
\IEEEoverridecommandlockouts                              

\usepackage{graphics}    
\usepackage{times}       
\usepackage{amsmath}     
\usepackage{amssymb}     
\usepackage{graphicx}
\usepackage{tabularx}
\usepackage{booktabs}
\usepackage{multirow}

\usepackage[separate-uncertainty=true,number-unit-product=\ ]{siunitx}%
\usepackage{subcaption}
\usepackage{tabularx}
\usepackage[protrusion=true,expansion]{microtype}
\usepackage{caption}
\usepackage[hidelinks]{hyperref} 
\usepackage{makecell}
\usepackage{flushend}

\def\eqref#1{Eq.~(\ref{#1})}

\newcommand\etal{\emph{et al. }}

\title{\LARGE \bf Demonstration-Enhanced Adaptable Multi-Objective Robot Navigation}

\author{Jorge de Heuvel \and  Tharun Sethuraman \and Maren Bennewitz
  \thanks{
   All authors are with the Humanoid Robots Lab, University of Bonn, Germany. M. Bennewitz and J. de Heuvel are additionally with the Lamarr Institute for Machine Learning and Artificial Intelligence and the Center for Robotics, Bonn, Germany. 
   This work has been partially funded by the BMBF, grant No. 16KIS1949 and within the Robotics Institute Germany, grant No. 16ME0999, as well as by the DFG, grant No. \mbox{BE~4420/2-2}~(FOR 2535 Anticipating Human Behavior).
   }
   \thanks{
   We sincerely thank Murad Dawood for his valuable feedback in revising the manuscript. 
   }
}

\begin{document}
\maketitle
\thispagestyle{empty} 
\pagestyle{empty}

\begin{abstract} 
Preference-aligned robot navigation in human environments is typically achieved through learning-based approaches, utilizing user feedback or demonstrations for personalization.
However, personal preferences are subject to change and might even be context-dependent.
Yet traditional reinforcement learning (RL) approaches with static reward functions often fall short in adapting to evolving user preferences, inevitably reflecting demonstrations once training is completed.
This paper introduces a structured framework that combines demonstration-based learning with multi-objective reinforcement learning (MORL).
To ensure real-world applicability, our approach allows for dynamic adaptation of the robot navigation policy to changing user preferences without retraining. 
It fluently modulates the amount of demonstration data reflection and other preference-related objectives.
Through rigorous evaluations, including a baseline comparison and sim-to-real transfer on two robots, we demonstrate our framework's capability to adapt to user preferences accurately while achieving high navigational performance in terms of collision avoidance and goal pursuance.
\end{abstract}

\section{Introduction}
\label{sec:intro}

Mobile robot navigation has significantly advanced with deep reinforcement learning (RL), enabling end-to-end policies that traverse complex environments with foresighted and nuanced behaviors. 
In scenarios involving human-robot interaction, however, it becomes crucial to align these policies with user preferences~\cite{vamplew_human-aligned_2018, de_heuvel_learning_2022}, e.g.,  on approaching behavior, proxemics, and navigational efficiency, to achieve acceptance~\cite{mavrogiannis_core_2023}.

However, traditional RL-based navigation methods typically optimize for static and pre-configured objectives in their reward scheme such as path efficiency or obstacle avoidance~\cite{perez-darpino_robot_2021}, neglecting user preferences and their variability over time.
As a result, these methods lack mechanisms to adapt to shifting user preferences dynamically and require retraining to accommodate behavior changes, highlighting a significant gap in the current methodology.

A common strategy for addressing user preferences is learning from demonstrations.
To preference-align RL-based navigation around the human, de Heuvel~\etal~\cite{de_heuvel_learning_2022, de_heuvel_learning_2023-1} have employed an additional behavior cloning loss driven by demonstration data.
However, these approaches do not provide principled ways to dynamically trade off demonstrated behaviors against core navigation objectives such as efficiency and collision avoidance. 
This can lead to overly conservative or inconsistent behavior, reducing usability in real-world applications.
It becomes essential to devise mechanisms that can modulate the influence of demonstrations by user preferences, even after training.

\begin{figure}[t]
	\centering
	\includegraphics[width=0.98\linewidth]{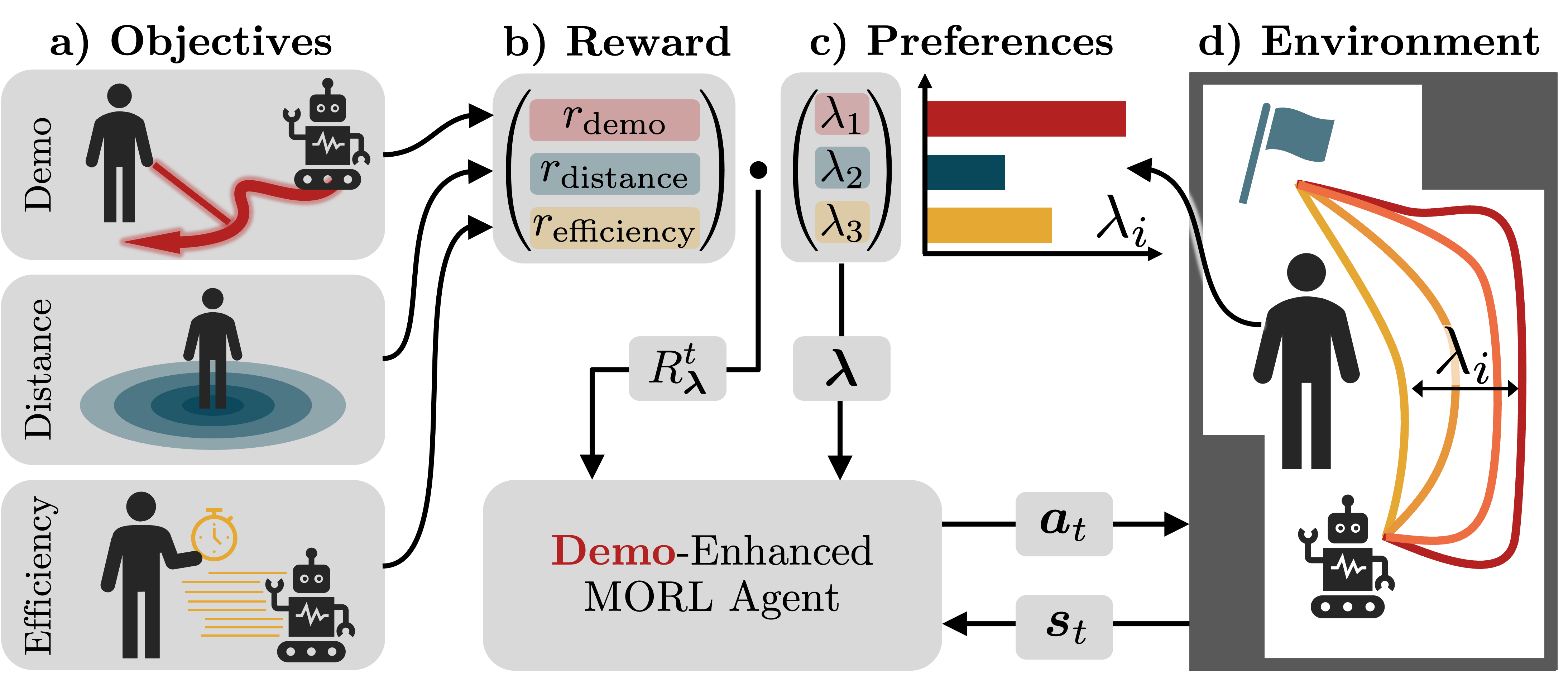}
	\caption{
		Our framework integrates demonstration-based learning into multi-objective reinforcement learning, enabling robots to adapt navigation policies to users' changing preferences even after training.
		\textbf{a)} The navigation style can fluently shift between demonstration-induced, distance keeping, and efficiency objectives.
		\textbf{b)} We modulate the MORL reward vector~$\boldsymbol{r}_t$ with a \textbf{c)} varying preference $\boldsymbol{\lambda}$, while providing $\boldsymbol{\lambda}$ as input to the agent.
		\textbf{d)} The resulting human-centered policy can generate a spectrum of trajectories, here sketched for the objectives of demonstration-reflection (red, here: wall-following) and path efficiency (yellow).
		\label{fig:motivation}}
\end{figure}

To overcome these challenges, we propose a novel framework that integrates demonstration-based learning (LfD) into multi-objective reinforcement learning~(MORL) to achieve flexible and preference-aware robot navigation (see Fig.~\ref{fig:motivation}). 
This combination extends MORL's on-the-fly policy adaptation capabilities~\cite{hayes_practical_2022} by modulating the influence of demonstrations and other objectives without retraining.

Specifically, our combined approach of LfD and MORL provides a structured way to incorporate user demonstrations as one of multiple competing objectives, enabling situationally adaptable trade-offs between demonstration adherence and navigational core objectives
Focusing on the robotic application, our experimental results demonstrate robust performance and accurate preference reflection for both a static and moving user.
Finally, a comprehensive sim-to-real transfer on two different robotic platforms further validates the feasibility and robustness of our method in human-centered navigation tasks.

In summary, the main contributions of our work are: 
\begin{itemize} 
	\item A multi-objective reinforcement learning human-aware robot navigation framework that enables policy adaptation to preferences post-training.
	\item The structured incorporation of demonstration data as a tuneable objective.
	\item Comparative navigation experiments in simulation validating demonstration modulation, behavior adaptation, robustness and generalization, concluded by a real-world transfer and evaluation on two different robots.
\end{itemize}

\section{Related Work}
\label{sec:related}
The concept of user-aware personalized navigation is gaining momentum, emphasizing robots that adapt their strategies based on individual user preferences.
Users can express preferences through ranking trajectory queries~\cite{choi_fast_2020, keselman_optimizing_2023} or providing demonstrations~\cite{de_heuvel_learning_2022, de_heuvel_learning_2023-1, hwang_promptable_2024}. 
Both feedback types can distill a preference-aligned navigation policy. 
While trajectory ranking can be used to extract user preferences~\cite{marta_aligning_2023}, this work establishes a demonstration-infused policy that aligns on-the-fly without retraining through multi-objective reinforcement learning (MORL).

The concept of optimizing for multiple objectives has already been applied in traditional non-RL navigation approaches~\cite{ferrer_anticipative_2019, kumar_modified_2023, banisetty_socially_2021}.
Traditional methods however are limited by their inability to integrate preference-conveying demonstration data.
In the context of RL, MORL extends standard RL by enabling the simultaneous optimization of multiple objectives.
MORL frameworks exist for discrete~\cite{yang_generalized_2019} and continuous action spaces~\cite{basaklar_pd-morl_2023, xu_prediction-guided_2020}, while the latter are particularly interesting for robotic tasks.
So far, MORL has been applied to autonomous driving~\cite{he_toward_2023} and robotic tasks such as manipulation~\cite{huang_constrained_2022}, navigation~\cite{marta_aligning_2023, cheng_multi-objective_2023, cheng_multi-objective_2024, lee_adaptive_2023-1}, and path planning~\cite{wilde_scalarizing_2024}.

\mbox{Ballou~\etal~\cite{ballou_variational_2023}} used meta reinforcement learning to adjust robot navigation among humans, efficiently fine-tuning policies for changes in the reward function, such as goal pursuance or distance keeping. 
However, their adaptation to shifting objectives is not instantaneous but rather requires an adaptation training phase. 
In contrast, our MORL policy adapts to preference weight changes in the preference space immediately.

Cheng~\etal~\cite{cheng_multi-objective_2023} proposed a MORL-based navigation policy that adapts to dynamic preferences over multiple navigation objectives in human environments, utilizing deep Q-networks for preference-weighted action selection. 
Similar to our approach, their method processes 2D~lidar data as input. 
However, unlike our approach, they employ a discrete action space with acceleration commands, whereas we utilize MORL-enabled TD3 actor-critic architecture with a continuous action space of linear and angular velocity control for smooth motions.

Cheng~\etal~\cite{cheng_multi-objective_2024} presented an approach to learn robot navigation in human-populated environments leveraging a multi-objective reward vector formulation.
Compared to our study, they are not accounting for different preferences, as their approach optimizes a fixed set of objectives without mechanisms to adjust trade-offs dynamically. 
Choi~\etal~\cite{choi_fast_2020} proposed to use multi-agent training with parameterized rewards and action commands for adaptable robot navigation. 
Parameterized rewards can be used with standard RL policies, potentially at the cost of weaker multi-objective optimization.
In contrast, our agent estimates Q-values for different objectives separately while incorporating tunable demonstrations alongside other navigation objectives. 

Hwang~\etal~\cite{hwang_promptable_2024} proposed a vision-based MORL framework for adapting robot navigation with discrete actions to human preferences through demonstrations, trajectory comparisons, and language instructions. 
However, their use of demonstrations is limited to estimating corresponding best-representing preference weights based on given objectives, possibly losing nuanced behavior traits in the demonstration data, whereas our approach directly integrates demonstration data to shape navigation behavior.

\section{Our Approach}
\subsection{Problem Statement}
We consider a wheeled robot navigating in the vicinity of a human and unknown obstacles, pursuing a local goal while avoiding collisions.
The robot is controlled via continuous velocity commands.
The human has certain preferences about the navigation style of the robot that may change depending on navigational context, such as task or time constraints, and which should be considered by the robot while navigating to the goal.
These navigation preferences can be expressed both in the form of a preference vector and demonstrations.
We assume the robot is provided a robot-centric goal location and can reliably estimate the human position, obstacles are perceived by the robot through 2D~lidar.
The navigation policy processes sensor data and goal information along with a preference vector containing user preferences, allowing for on-the-fly behavior adaptation within a single policy.
Our approach explicitly focuses on single-human interaction, personalizing robot behavior based on individual user preferences rather than group dynamics.
Code for our approach is available online.\footnote{Code repository: \url{https://github.com/HumanoidsBonn/demo_enhanced_morl_nav}}

\subsection{Multi-Objective Reinforcement Learning}
\label{sec:pd_morl}
Multi-objective reinforcement learning (MORL) enhances traditional RL by integrating multiple, often conflicting, objectives~\cite{hayes_practical_2022}.
In MORL, the agent is trained to learn policies that strike a balance among these diverse objectives, as opposed to a one-dimensional reward function.
The MORL problem is formulated within the framework of a Markov Decision Process (MDP), defined by the tuple $(\mathcal{S}, \mathcal{A}, \mathcal{P}, \mathcal{R}, \gamma)$. 
Here, $\mathcal{S}$ is the state space, $\mathcal{A}$ is the action space, $\mathcal{P}: \mathcal{S} \times \mathcal{A} \times \mathcal{S} \rightarrow [0,1]$ is the state transition probability, and $\gamma$ is the discount factor. 
A distinctive feature of MORL is the multi-dimensional reward function \(\mathcal{R}: \mathcal{S} \times \mathcal{A} \rightarrow \mathbb{R}^n\), which outputs a vector of rewards \(\boldsymbol{r}_t\) for \(n\) different objectives.

A single policy optimally adheres to a given combination of preferences, represented by the convex preference weight vector \(\boldsymbol{\lambda} \in \mathbb{R}^n\).
The learning algorithm optimizes a scalarized reward function \(R_{\boldsymbol{\lambda}}(s, a) = \boldsymbol{\lambda}^\top \boldsymbol{r}(s, a)\), itemizing the different objectives.

We employ the preference-driven (PD-)MORL implementation of Basaklar~\etal~\cite{basaklar_pd-morl_2023}, precisely their TD3-based algorithm, which can learn a single-network policy that covers the entire preference space. 
PD-MORL achieves this by four major modifications to TD3's standard actor-critic-structure with respect to the policy loss and preference-space exploration:
i) A preference interpolator $I(\boldsymbol{\lambda})=\boldsymbol{\lambda}_p$ projects the original preference vectors $\boldsymbol{\lambda}$ into a normalized solution space, thereby improving the aligning of preferences with multi-objective value solutions $\boldsymbol{Q}$. 
ii) The framework is complemented by an angle loss $g(\boldsymbol{\lambda}_p, \boldsymbol{Q})$, designed to minimize the directional angle between the interpolated preference vectors $\boldsymbol{\lambda}_p$ and the multi-objective vector $\boldsymbol{Q}$, thus improving preference-reflection.
The actor network is updated by maximizing the term $\boldsymbol{\lambda}^T \boldsymbol{Q}$, where $\boldsymbol{\lambda}$ is the original convex preference vector and $\boldsymbol{Q}$ is the critic network's vector-based output, while simultaneously minimizing the directional angle term. 
iii) To efficiently learn across the entire preference space in PD-MORL, a hindsight experience replay mechanism~\cite{andrychowicz_hindsight_2017} enhances the preference vector diversity during training.
iv) The training process involves running a number of \(C_p\) environments in parallel for \(N\) time steps, each tailored to explore a distinct segment of the preference-vector space.

While Basaklar~\etal originally evaluated their TD3-based PD-MORL on gym benchmarks~\cite{xu_prediction-guided_2020}, we extend it to a three-objective robotic navigation task.
The focus of our study is on task-related behavior adaptability, robustness, generalization, and real-world deployment performance.
To the best of our knowledge, our study represents the first application of the PD-MORL framework to real-world robot tasks, where sensor-induced noise and partial observability introduce additional challenges.

\subsubsection{State and Action Space}
The state space includes the local goal, human position, and obstacles detected by a lidar sensor. 
The agent receives the relative 2D~goal location $\boldsymbol{p}_g$ and human position $\boldsymbol{p}_h$ in polar coordinates. 
The $360^\circ$ lidar scan, with a range of $4 \si{\meter}$, is min-pooled from 720 to $N_\text{lidar} = 30$ rays. 
These are combined in the state vector as $s_t = (\boldsymbol{p}_g, \boldsymbol{p}_h, \mathcal{L}_t)$, where \mbox{$\mathcal{L}_t = { {d}_i^t | 0 \leq i < N_\text{lidar} }$}.

The robot is controlled with linear and angular velocity commands $a_t = (v, \omega)$, where $v \in [0, 0.5] \si{\meter\per\second}$ and $\omega \in [-\pi, \pi] \si{\radian\per\second}$.
The perception-action loop runs at $5 \si{\hertz}$.

\subsubsection{Networks}
The networks of actor, critic, behavior cloning policy, and reward model (see below) are fully connected multi-layer perceptron (MLP) networks with an identical architecture consisting of 4 layers with 256 neurons each.
The uniform architecture is a heuristic choice, validated in preliminary experiments.

\subsection{Incorporating Demonstrations}
\begin{figure}[t]
	\centering
	\includegraphics[width=0.95\linewidth]{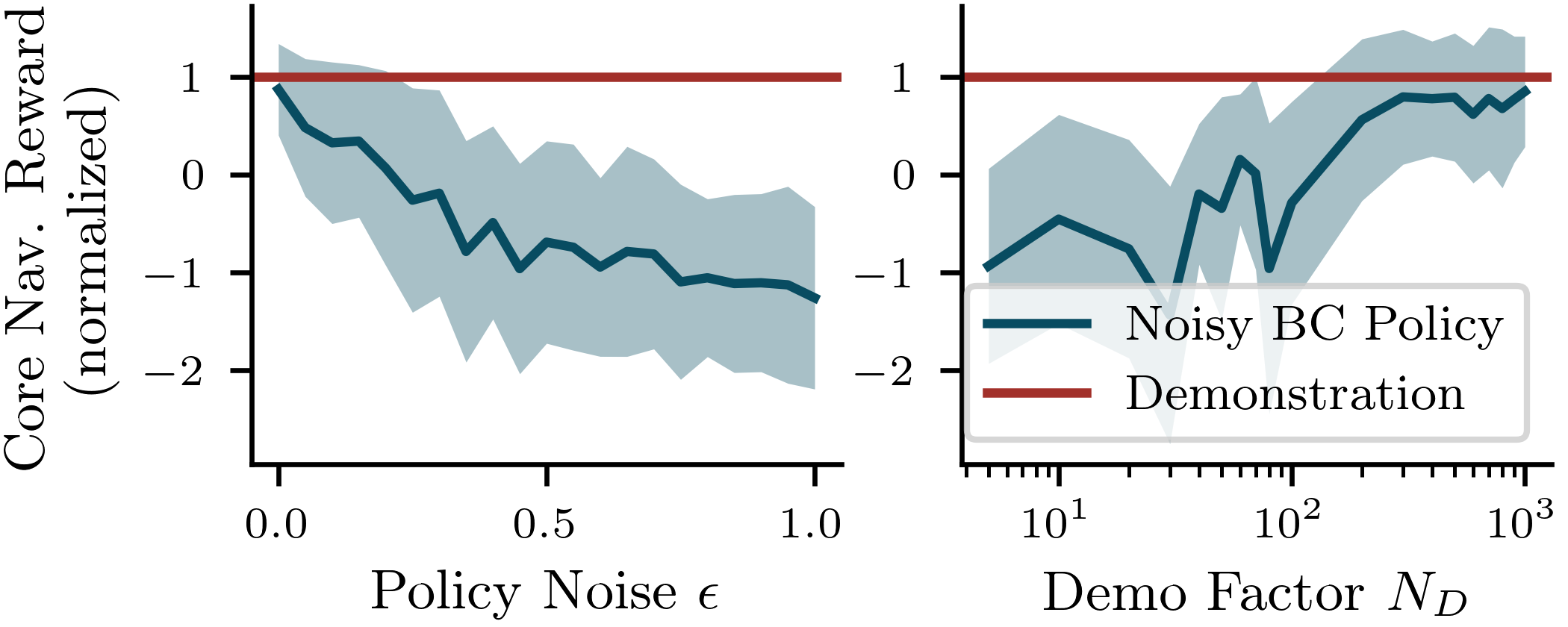}
	\caption{
		Exploration of D-REX-related demonstration parameters averaged over 20 trajectory rollouts, measured against the optimal demonstration behavior's reward.
		\textbf{a)} The execution of the $\epsilon$-greedy noise-injected behavior cloning (BC) policy trained with a demonstration augmentation factor of $N_D=$ 1,000 reveals a degradation of navigation performance measured by the normalized core reward $r_\text{core}$ with growing strength of the injected noise.
		\textbf{b)} The demonstration augmentation factor $N_D$ indicates how many times the optimal human-centric demonstration trajectory (see Sec.\ref{sec:demo_reward}) was rolled out with randomized obstacle placement to form the training dataset, showing increased performance with higher $N_D$.
		\label{fig:drex_demonstration}}
\end{figure}
\label{sec:demos}
As one of our main contributions, we distill nuanced navigation from demonstration trajectories $\tau$ into a reward model that natively integrates into MORL as one of the objectives and guides the learning agent to demonstration-like behavior.
Through this novel design choice, the influence of demonstrations can be modulated by $\mathbf{\lambda}$ post-training.

A reward model is typically derived from pairwise A$\succ$B preference queries in a human feedback process via a ranking loss~\cite{marta_aligning_2023}.
However, demonstrations are typically considered equally important, rendering them unsuitable for a ranking-based reward model.
Addressing the problem of non-existent ranking from demonstration data, we use a workaround involving artificial rankings. 
We employ the disturbance-based reward extrapolation (D-REX) approach by \mbox{Brown~\etal~\cite{brown_better-than-demonstrator_2020}}, which imitates pairwise A$\succ$B preference queries by ranking over noise-injected demonstration trajectories.
First, a behavior cloning~(BC) policy $\pi_{BC}$ is trained from $N_D$ demonstration trajectories, the collection of which is outlined in Sec.~\ref{sec:demo_reward}.
Subsequently, the BC policy $\pi_{BC}(\cdot|\epsilon)$ is executed with increasing level of $\epsilon$-greedy policy noise $\epsilon \in \mathcal{E} =(\epsilon_1, \epsilon_2, \ldots, \epsilon_d)$ with \mbox{$\epsilon_1 < \epsilon_2 < \ldots < \epsilon_d$}.
In short, low-noise trajectories almost perfectly resemble the demonstration trajectory, while they slowly lose their shape with growing levels of noise.
Trajectory rollouts generated with lower noise are automatically ranked superior compared to their higher-noise counterparts.
Finally, a rich preference-ranking dataset 
\begin{equation*}
	D_{\text{rank}} = \{\tau_i \prec \tau_j | \tau_i \sim \pi_{BC}(\cdot|\epsilon_i), \tau_j \sim \pi_{BC}(\cdot|\epsilon_j), \epsilon_i > \epsilon_j \}
\end{equation*}
is obtained.
From $D_{\text{rank}}$, we train a reward model $\hat{R}(s,a) \in [0, 1]$ using the Bradley-Terry model~\cite{bradley_rank_1952} with its typical implementation as a binary cross entropy loss such that $\sum_{s \in \tau_i} \hat{R}_\theta(s,a) < \sum_{s \in \tau_j} \hat{R}_\theta(s,a) \text{ when } \tau_i \prec \tau_j$.

For our ranking dataset $D_{\text{rank}}$, we choose a noise range $\mathcal{E} =(\num{0}, \ldots, \num{0.2})$ and obtain $N_D=$~1,000 demonstration augmentations with obstacle randomization from a single demonstration pattern.

\subsection{Reward Vector}
The reward vector covers traditional navigational objectives, subsequently referred to as core objectives, and three tuneable distinct style objectives based on quantifiable metrics and preference demonstrations.
In our MORL setup, the core objectives are summed and occupy the first entry in the reward vector $\boldsymbol{r}_t$ which is assigned a static preference weight of one.
Note that this is neglected in further notations of the convex vector $\boldsymbol{\lambda}$ to focus on the tuneable objectives.
For the other objectives occupying entries in the reward vector, the preference weights are dynamic.
The reward vector for our MORL framework consists of four components as explained below:
\begin{align}
	\boldsymbol{r}_t = (\underbrace{r_\text{core}^t}_\text{static},  \underbrace{r^t_\text{demo}, r_\text{distance}^t, r^t_\text{efficiency}}_\text{dynamic objectives})
\end{align}

\subsubsection{Navigational Core Objectives}
Independent of preferences, the agent must exhibit goal pursuance and collision avoidance. 
Goal-oriented navigation is achieved by a continuous reward $r_\text{goal}^t = 125 \cdot (d_g^t - d_g^{t-1})$, based on the change in distance $d_g = |\boldsymbol{p}_g|$ from the goal. 
The total cumulative goal reward $R = \sum_{t=0}^{T} r_\text{goal}^t$ is non-discounted to remain independent of the number of steps to the goal, avoiding a bias towards shortest paths and thus the efficiency preference objective. 
Collision avoidance uses a sparse penalty $r^t_\text{collision} = -$1,000 for contact between the robot and any obstacle. 
The core reward function is $r_\text{core}^t = r_\text{goal}^t + r^t_\text{collision}$.

\subsubsection{Tuneable Preference Objectives}
\begin{figure*}[!t]
	\includegraphics[width=0.99\linewidth]{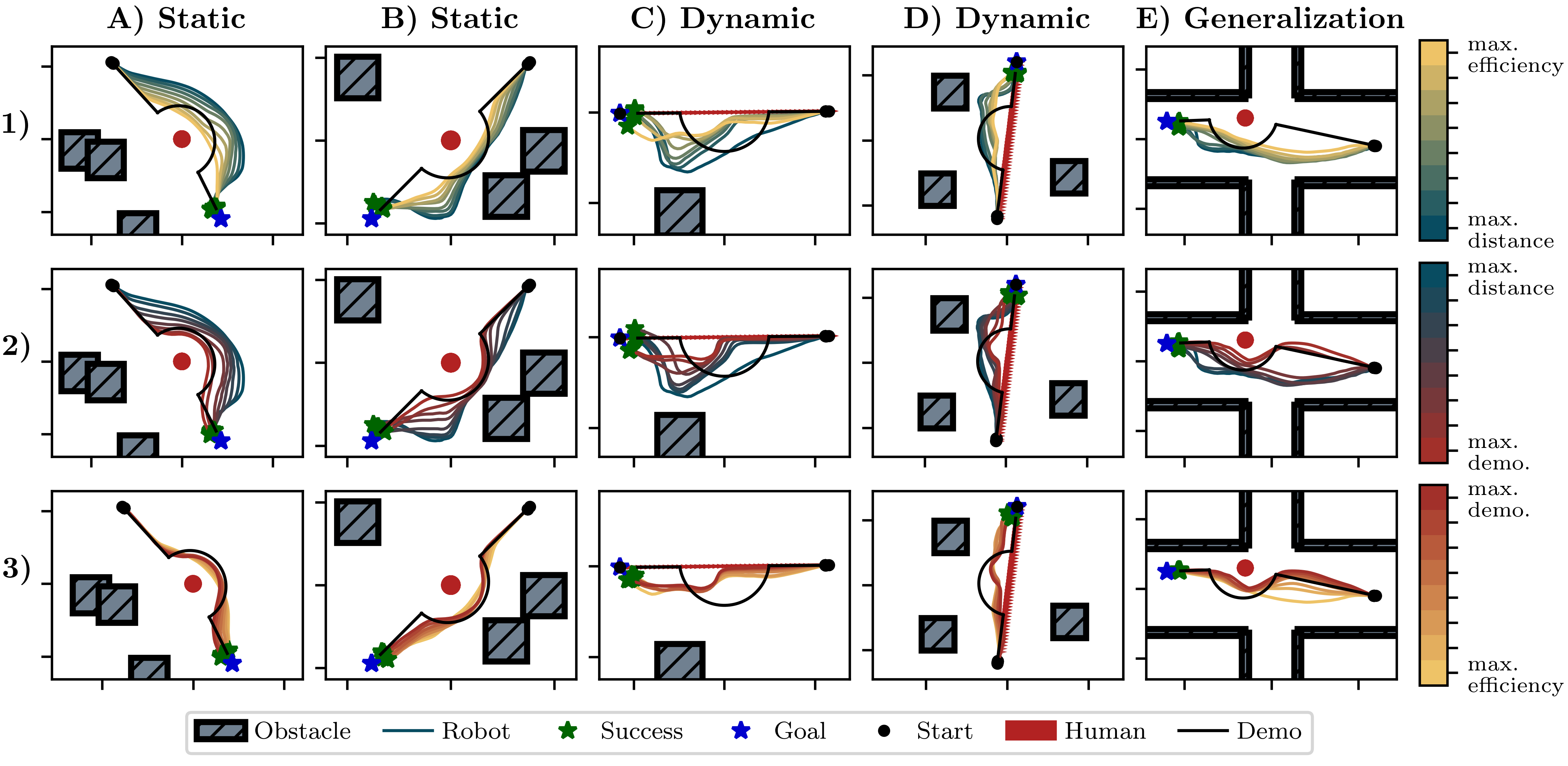}
	\centering
	\caption{
		Trajectory rollouts in simulation for different preference vectors \textbf{(rows)} and different scenes with  a static and a dynamic approaching human \textbf{(columns)}.
		As can be seen, the navigation policy shifts its behavior according to the set preference.
		The colorbars on the right indicate the interpolated preference space~$\Lambda_i$ for each plot row.
        Static scenarios such as \textbf{(A+B)} were covered during training, while a moving human \textbf{(C+D)} and the corridor environment \textbf{(E)} test for generalization.
		While shifting \textbf{Row 1)} from shortest driving behavior under the maximum efficiency preference (yellow) to distance-keeping (blue), the minimum distance from the human increases.
		At the same time, a tendency to navigate alongside obstacles - if present close to the path - has developed.
		Shifting towards the maximum demonstration preference \textbf{(Row 2)}, the trajectory shapes increasingly resemble the demonstration pattern (black).
		On the shift back to maximum efficiency \textbf{(Row 3)}, the demonstration pattern disappears in favor of shortest trajectories.
        Comparing the static \textbf{(A+B)} vs. moving human \textbf{(C+D)}, the demonstration preference reflection becomes less distinct as the agent struggles to follow the static pattern that moves with the now dynamic human, yet efficiency and distance preferences keep up with a moving human.
        In the corridor intersection scene \textbf{(E)}, not included during training of the policy, the agent successfully accounts for the wall, reducing the possible distance-keeping to the human.
        The varied angle between human and goal from the robot's perspective does not prevent the policy from first approaching the human under the maximum demonstration preference, before continuing towards the goal.
	}
	\label{fig:qualitative_enhanced}
\end{figure*}
Our three user-centric style objectives cover demonstration-reflection, efficiency, and proxemics:
To include proxemics, an important comfort factor in human-aware navigation, we define a quadratic distance penalty for positional closeness $d_h = |\boldsymbol{p}_h|$ to the human within a range $d_\text{thresh} = 2 \si{\meter}$ as
\begin{equation}
	r_\text{distance} = -10 \frac{(d_h - d_\text{thresh})^2}{(d_\text{thresh} - d_\text{min})^2} \text{ if } d_h \leq d_\text{thresh} \text{,}
\end{equation}
else zero, with $d_\text{min} = 0.3 \si{\meter}$. 

The second style objective is navigational efficiency, or shortest path navigation, implemented with a constant time penalty $r^t_\text{efficiency} = -10$.


The third and last objective is demonstration-like behavior $r^t_\text{demo}$, as elaborated below.
Note that all rewards of the tuneable objectives are defined as penalties with uniform range of $[-10, 0]$.

\subsubsection{Demonstration Acquisition and Reward}
\label{sec:demo_reward}
Demonstrations can capture nuanced navigation styles that are difficult to express using analytical reward functions, such as characteristically-shaped trajectories when approaching the user.
In this work, we rely on a predefined optimal demonstration pattern, see Fig.~\ref{fig:qualitative_enhanced}.A1 (black line), where the robot circumnavigates the human in a distinct circular manner. 
After directly approaching the human, at $d_h = 1 \si{\meter}$, it executes a $90^\circ$ left-hand turn and orbits the human clockwise at a radius $d_h$.
Once between human and goal, it turns left and proceeds directly towards the target.
While not being user-demonstrations, the distinct pattern enables a clear performance analysis, as its behavior is by design contradictory to the other two objectives, efficiency and distance-keeping.
Specifically, the trajectories are only partially goal-directed, conflicting $r^t_\text{efficiency}$, and traverse close to the human at $d_h = 1 \si{\meter}$, contradicting $r_\text{distance}^t$ with an impact radius of $2 \si{\meter}$.
Anchored solely around the human and the goal position, we can easily augment the single demonstration trajectory by rolling it out $N_D$ times in randomized obstacle configurations, recording only collision-free rollouts.
The resulting dataset is handed to the D-REX pipeline, as elaborated in \mbox{Sec.~\ref{sec:demos}}.
The final reward term is \mbox{$r^t_\text{demo} = -10 \cdot (\hat{R}_\theta(s_t,a_t) - 1)$}.

\section{Experimental Evaluation}
\label{sec:exp}
Our experimental evaluation is conducted to validate the following claims:
\begin{itemize}
	\item C1: The D-REX-based reward model successfully captures and teaches the demonstration patterns to the agent.
	\item C2: We learn a preference-adaptable, demonstration-modulating, yet reliable navigation policy.
    \item C3: PD-MORL is crucial to successfully learn our robot navigation task.
	\item C4: Our policy generalizes from simulation to the real world, even on a robot not used for training.
\end{itemize}
Our evaluation concludes with a sim-to-real transfer and evaluation on two robots.

\subsection{Training and Environment}
We train using the iGibson simulator~\cite{li_igibson_2022} with a simulated Kobuki TurtleBot 2. 
Robot start and goal positions are randomly sampled, 6 to $12 \si{\meter}$ apart in open space. 
A static human is placed between them, aligning with a static-human demonstration pattern.
Three static rectangular obstacles are randomly placed, avoiding occupied positions. 
The robot must navigate to the goal while avoiding both the human and obstacles, which may conflict with the human distance-keeping objective.
An episode terminates upon successfully reaching the goal, robot collision, or a timeout after 300~steps.
Training is conducted for 600k steps across $C_p = 3$ environments, using $\gamma = 1.0$, and the final model is used for evaluation.
For the evaluation of generalization to dynamic environments only, not training, we simulate a moving human approaching the robot with an opposite start goal configuration. 

\subsection{Qualitative Navigation Analysis}
\label{sec:qualitative_analyis}
Figure~\ref{fig:qualitative_enhanced} shows navigation strategies of our MORL agent in static (A+B+E) and dynamic human (C+D) scenarios in simulation, under varying preference weights and obstacle configurations. 
Three subplot rows interpolate convex preferences between pairwise combinations of two objectives, with the third objective fixed at zero.
In Row 1, preferences interpolate between distance and efficiency, parameterized by $\mu \in [0, 1]$, with the vector $\boldsymbol{\lambda}_1(\mu) = (0, \mu, 1-\mu)$. 
The other rows follow similar pairwise combinations. 
The resulting set of $\boldsymbol{\lambda}_i(\mu)$ is $\Lambda_i = \left\{ \left(\frac{i}{N}, 1 - \frac{i}{N}, 0\right) \mid \mu = \frac{i}{N}, i = 0, \ldots, N \right\}$ with $N=10$, forming the test set $\Lambda = \Lambda_1 \cup \Lambda_2 \cup \Lambda_3$ with a total of $33$~preference vectors, see Sec.~\ref{sec:quantitative_analyis}.

The plots depict the robot’s trajectories from an initial point (black dot) to a goal (blue star), considering static obstacles and a human (red circle \& arrow), with the optimal demonstration trajectory (black line) included.

Starting with the static human in Fig.\ref{fig:qualitative_enhanced}A+B, the shift from efficiency to distance-keeping (Fig.\ref{fig:qualitative_enhanced}.1) shows increasing human distance along the path, with the robot eventually passing closely without collision, reducing path length due to the efficiency penalty $r^t_{\text{efficiency}}$. 
Under maximum human distance preference, the robot occasionally stays close to obstacles before turning towards the goal after passing them.

For the shift from distance-keeping to demonstration-like behavior (Fig.~\ref{fig:qualitative_enhanced}.2), the minimum distance to the human decreases. 
Supporting C1, trajectories shape into the characteristic demonstration pattern of straight approach, circular circumnavigation, and a goal-directed turn, yet sharp corners near the human are less pronounced than in the demonstration.

Finally, shifting preferences from demonstration back to efficiency (Fig.~\ref{fig:qualitative_enhanced}.3), demonstration-driven trajectories bend around the human, while efficiency-driven ones head directly to the goal after passing. 
When obstacles are near the human, collisions are avoided, though at reduced distance. 
Under maximum distance preference, human distance is maintained before and after obstacles, and all trajectories pass the human on the right, following the demonstration pattern.

To further evaluate the generalization and robustness of our policy, we test it in a moving human environment and a previously unseen scene.
In this dynamic setting, which was not covered during training, a human approaches at $0.5 \si{\meter\per\second}$ (Fig.~\ref{fig:qualitative_enhanced}.C+D), the efficiency and distance-keeping objectives are maintained without collisions. 
The avoidance maneuvers occur more abruptly than in the static case, bending sharply away from the human. 
As expected, the demonstration pattern is less followed, with the orbiting part shrinking or not completed due to the moving human.

Similarly, we assess generalization and robustness in an unseen corridor intersection scenario (Fig.~\ref{fig:qualitative_enhanced}.E).
The agent successfully accounts for the presence of the wall, which limits the possible distance it can maintain from the human.
Despite the varied angle between the human and the goal from the robot’s perspective, the policy prioritizes initial approach behavior, aligning with the maximum demonstration preference, before continuing toward the goal. 
This indicates that the learned policy generalizes to unseen spatial configurations while adhering to key objectives.

These results provide evidence for C1 and C2, showing the robot’s ability to adjust its behavior from human-distant to demonstration-driven and efficiency-focused navigation.

\subsection{Quantitative Analysis}
\label{sec:quantitative_analyis}
\begin{figure}[t]
	\centering
	\includegraphics[width=0.999\linewidth]{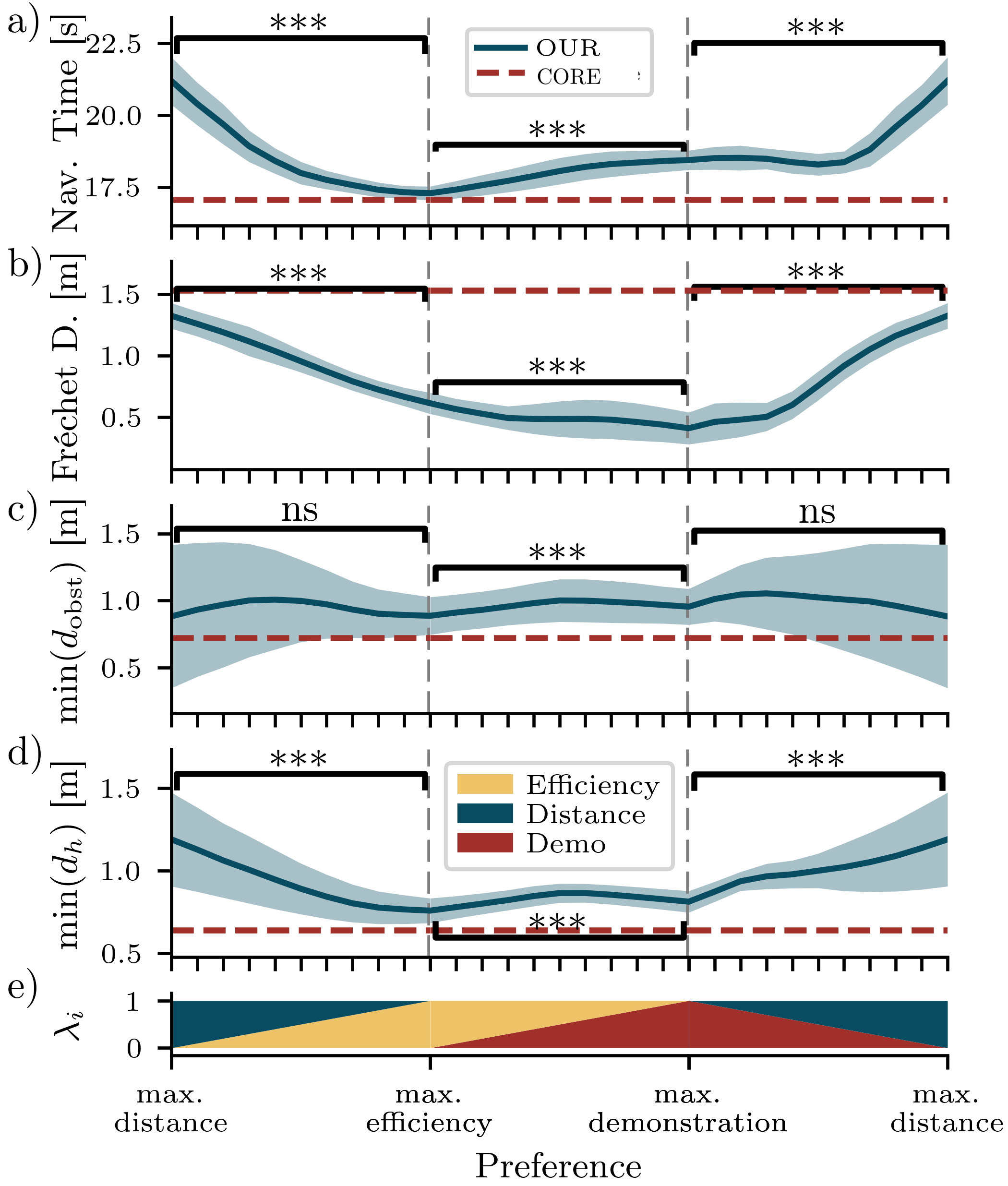}
	\caption{
		Quantitative metrics of OUR agent for different preference configurations \textbf{(e)}, tested for statistical significance for dissimilar means between the maximum preferences, with *** for $p<.001$, and ns for not significant.
		\textbf{a)} The navigation time is smallest for maximized efficiency preference, as expected.
		\textbf{b)} The Fréchet distance to the demonstration trajectory decreases as the demonstration preference increases.
		\textbf{c)} The minimum distance to any obstacle is measured using the lidar.
		\textbf{d)} The minimum distance from the human grows with its preference weight.
		The preference-independent non-MORL policy CORE ~(red dotted line) that only obeys the navigational core reward term $r_\text{core}$ of collision avoidance and goal pursuance is included in each plot.
		\label{fig:quantitative}}
\end{figure}
\subsubsection{Preference Reflection}
We conducted a quantitative evaluation of the preference-reflecting agent using multiple performance and navigation metrics (Fig.~\ref{fig:quantitative}). 
The agent was tested across 100 episodes in random environments, using different interpolated preference weights $\boldsymbol{\lambda} \in \Lambda$ (colored fractions in Fig.\ref{fig:quantitative}e; see Sec.~\ref{sec:qualitative_analyis}). 
Statistical significance between mean values for the maximum preferences was assessed using a Student’s t-test with Bonferroni-correction.

The agent (OUR) achieved a success rate of $\num{100} \si{\percent}$ with no timeouts or collisions (Table~\ref{tab:ablation}, first column). 
As the distance preference increases, both minimum human distance and navigation time rise (Fig.~\ref{fig:quantitative}a+d), indicating longer trajectories to maintain greater human distance.

To assess how well the demonstration trajectory is reflected (claim C1), we computed the Fréchet distance~\cite{alt_computing_1995} between the demonstration and executed trajectories (Fig.~\ref{fig:quantitative}b). 
The minimum mean Fréchet distance of $\num{0.41} \si{\meter}$ occurs when demonstration preference is maximized. 
Efficiency and distance-keeping preferences also reduce the Fréchet distance, as the demonstration path passes close to the human.

Comparing the trends of minimum obstacle distance ($\min(d_\text{obst})$, Fig.\ref{fig:quantitative}c) and minimum human distance ($\min(d_h)$, Fig.\ref{fig:quantitative}d), the agent clearly distinguishes between humans and static obstacles. 
As the human distance preference increases, the robot maintains a larger distance from the human, while staying close to obstacles, accepting higher collision risk to prioritize proxemic preferences.

Our quantitative analysis supports the findings from the qualitative evaluation, providing measurable evidence for research claims C1 and C2.

\setlength\lightrulewidth{0.1ex}
\begin{table}[b!]
	\centering
    \setlength{\tabcolsep}{3pt} 
	\begin{tabular}{rccccccc}
		 Metric & $\boldsymbol{\lambda}$ & OUR & -NH & -RM & -RM-NH & SAC-PR & -PR-$\gamma$\\
		\toprule
		SR$\uparrow$ [\%] & $\Lambda$ & \textbf{100} & 96.8 & 100 & 79.6 & 45.4 & 54.5\\
		CR$\downarrow$ [\%] & $\Lambda$ & \textbf{0} & 2.7 & 0 & 11.4 & 53.2 & 44.4 \\
		TR$\downarrow$ [\%] & $\Lambda$ & \textbf{0} & 0.5 & 0 & 9.0 & 1.2 & 1.1 \\
		\midrule[\lightrulewidth]
		$\min(d_h)$$\uparrow$  [m] & $\boldsymbol{\lambda}_\text{dist}$ & 1.18 & 0.52 & 1.16 & 0.48 & 1.06 &  0.91\\
		Fréchet$\downarrow$  [m] & $\boldsymbol{\lambda}_\text{demo}$ & \textbf{0.41} & 0.57 & 0.46 & 0.49 & - &  1.06 \\
		Nav. time$\downarrow$ [s] & $\boldsymbol{\lambda}_\text{eff}$ & 17.3 & \textbf{16.9} & 17.4 & 19.2 & - &  20.8 \\
		\bottomrule
	\end{tabular}
	\caption{
		Quantitative analysis, ablation, and baseline study with respect to the state space and reward model, bold number highlighting the highest performance.
		For the ablation identifiers and preference vectors $\{\boldsymbol{\lambda}_\text{dist}, \boldsymbol{\lambda}_\text{demo}, \boldsymbol{\lambda}_\text{eff}\}$, please refer to Sec.~\ref{sec:ablation}.
		For brevity, the identifiers are shortened after OUR, so that, e.g., -NH corresponds to OUR-NH with the human pose state excluded.
        The baselines with parameterized rewards are denoted with SAC-RP and SAC-PR-$\gamma$, short -PR-$\gamma$.
		The results were averaged over 100~trajectories for single $\boldsymbol{\lambda}$, and for the success rate~(SR), collision rate~(CR), and timeout rate~(TR) additionally over all $\boldsymbol{\lambda}_i \in \Lambda$, precisely \mbox{$33 \times 100 = 3,300$} trajectories.
        The baseline SAC-PR had no successful trajectories under $\boldsymbol{\lambda}_\text{demo}$ and $\boldsymbol{\lambda}_\text{eff}$.
	}
	
	\label{tab:ablation}
\end{table}

\subsubsection{Ablation Study}
\label{sec:ablation}
We ablated the architecture with respect to the state space and demonstration reward model, compare Table~\ref{tab:ablation}.
The state space changes apply to all involved models: D-REX BC policy, D-REX reward model, actor, and critic.
The ablations cover exclusion of human position (OUR-NH), removal of the action $a_t$ as input to the reward model leaving $r^t_\text{demo} = \hat{R}_\theta(s_t)$ (OUR-RM), and the combination of both (OUR-RM-NH).
Note that the maximum preference vectors in Table~\ref{tab:ablation} are $\boldsymbol{\lambda}_\text{demo}=(1,0,0)$, $\boldsymbol{\lambda}_\text{dist} = (0,1,0)$, $\boldsymbol{\lambda}_\text{eff}=(0,0,1)$, respectively.

Compared to OUR, removing the human position from the state space in OUR-NH and OUR-RM-NH reduces distance-reflection capabilities. 
This is expected due to the correlation between human position and distance preferences in demonstrations.
While OUR-RM performs with a similar collision rate, its preference-reflection is slightly weaker than OUR. 

\subsection{MORL Baseline}
As single-policy MORL approaches with continuous action spaces are scarce due to the novelty of Basaklar's PD-MORL TD3-based algorithm, we implement an equivalent actor-critic-based MORL baseline with parameterized reward (-PR), analogous to the baselines in~\cite{he_toward_2023}.
Specifically, the handling of reward and $Q$ value differs:
In the baseline, the critic predicts a scalar $Q$ corresponding to the parameterized reward function \(R_{\boldsymbol{\lambda}}(s, a) = \boldsymbol{\lambda}^\top \boldsymbol{r}(s, a)\), thereby learning a mixed representation of all objectives. 
In contrast, PD-MORL's critic outputs a vector-based $\boldsymbol{Q}$, with each component corresponding to a separate objective, thus maintaining objective-specific representations throughout learning.
Furthermore, the four performance-boosting modifications of PD-MORL are not included in the baseline, compare Sec.~\ref{sec:pd_morl}.
The learning task characteristics and reward vector remain unchanged.

During training, convex preference weights are sampled at the beginning of each episode. 
Among the baseline actor-critic implementations, TD3 failed to converge on the task, whereas SAC~\cite{haarnoja_soft_2018-1} achieved better results. 
Performance further improved when adjusting the discount factor from $\gamma = 1.0$ to $\gamma$ = 0.98 in SAC-PR-$\gamma$ (see Tab.~\ref{tab:ablation}). 
Nevertheless, both SAC-PR and SAC-PR-$\gamma$ average in success below \SI{55}{\percent}.
Note that SAC-PR and SAC-PR-$\gamma$ show weaker preference reflection as compared to OUR, while SAC-PR failed entirely on the edge-case preferences $\boldsymbol{\lambda}_\text{demo}=(1,0,0)$ and $\boldsymbol{\lambda}_\text{eff}=(0,0,1)$.
The results highlight the superiority of PD-MORL for learning the robot navigation task, supporting C4.

\subsubsection{Non-MORL Core Navigation Agent}
To contextualize the core navigation objectives, we train and quantitatively evaluate a preference-independent, non-MORL policy CORE that optimizes only the navigational core rewards $r_\text{core}$ (goal and collision), compare the red dotted line in Fig.~\ref{fig:quantitative}.
Two metrics stand out:
The MORL agent prioritizes obstacles over humans, while the non-MORL baseline, lacking a human-distance reward, treats both similarly. 
This results in comparable minimum values (\( d_h = \num{0.64} \si{\meter}, d_\text{obst} = \num{0.72} \si{\meter} \)), contrasting with our MORL agent. Its higher demonstration Fréchet distance further confirms the absence of demonstration knowledge.

\subsection{Real-World Transfer}
\label{sec:real_world_analysis}
\begin{figure}[t]
	\centering
	
	\begin{subfigure}[b]{0.75\linewidth}
		\centering
		\includegraphics[width=\linewidth]{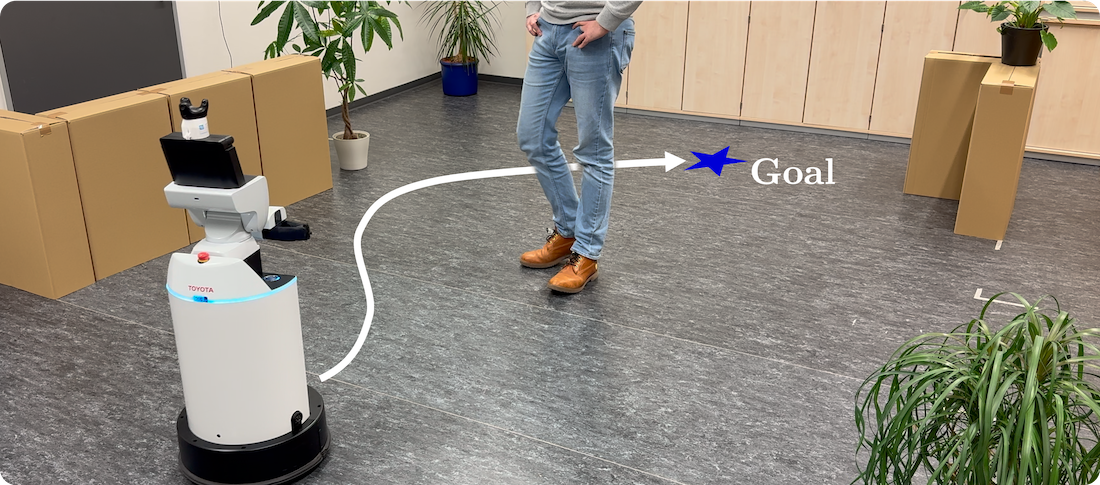}
	\end{subfigure}\\ 
	\vspace*{0.2em}
	\begin{subfigure}[b]{0.95\linewidth}
		\centering
		\includegraphics[width=\linewidth]{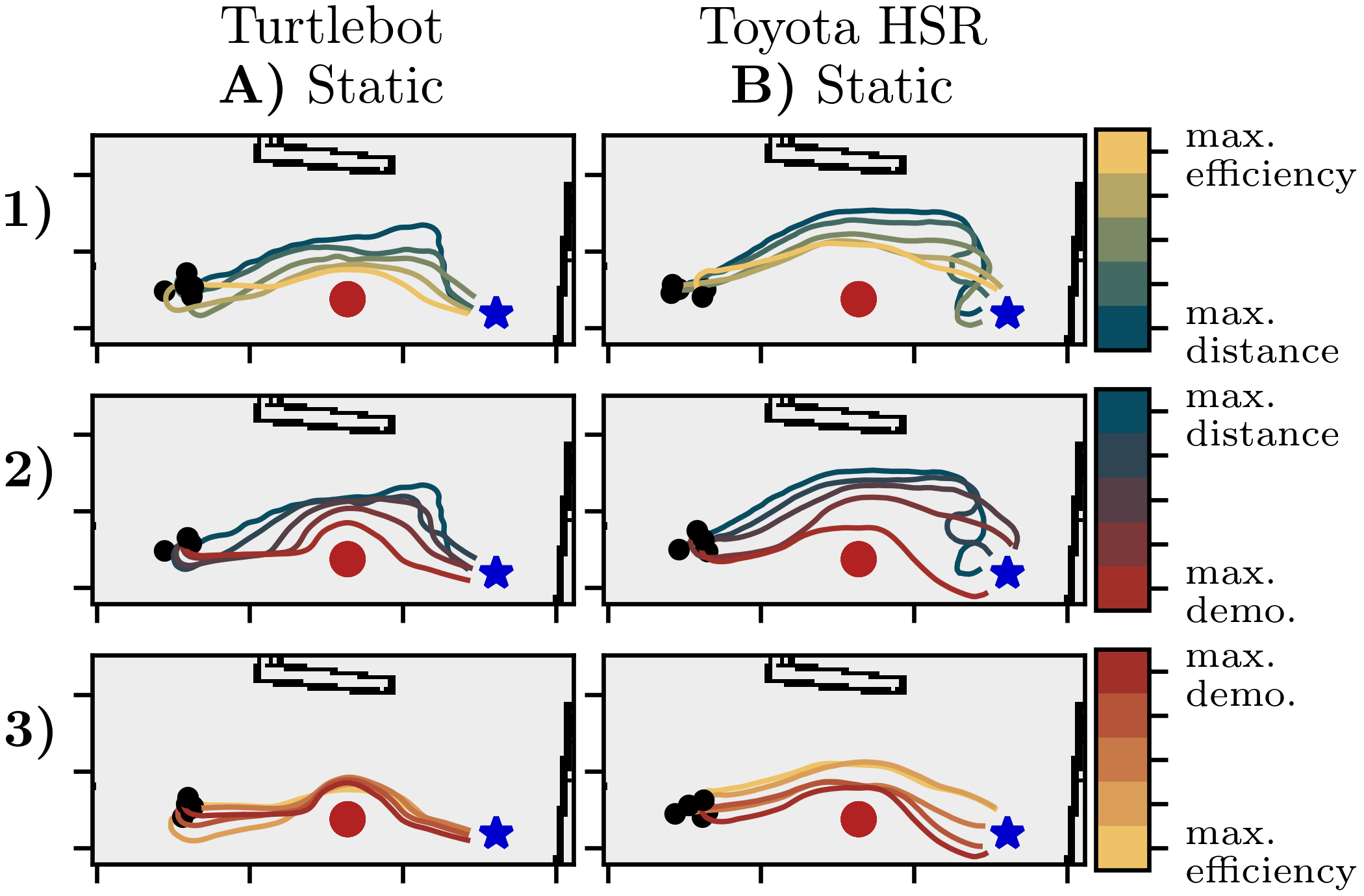}
	\end{subfigure}
     \vspace*{0.2em}
	\begin{subfigure}[b]{0.95\linewidth}
		\centering
		\includegraphics[width=\linewidth]{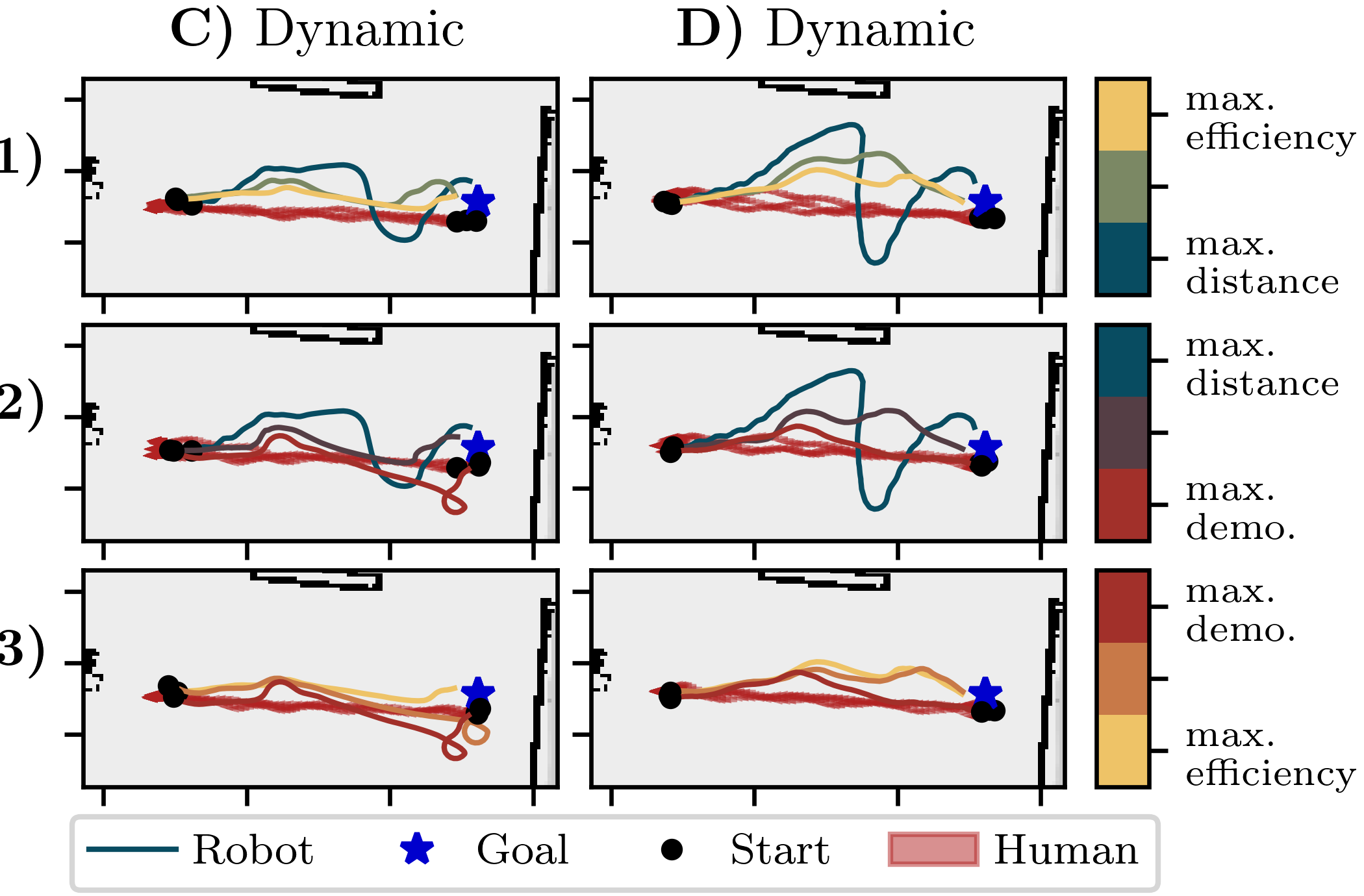}
	\end{subfigure}
	\caption{Real-world experiment setup \textbf{(top)} and results \textbf{(bottom)} with the policy OUR in a  sim-to-real transfer with the Kobuki TurtleBot 2 \textbf{(left)} and a the Toyota HSR \textbf{(right)}.
	With a static human as during training~\textbf{(A+B)}, the navigation behavior in the real world successfully reflects varying the preferences on both robots.
	While the TurtleBot exhibits better demonstration reflection, the HSR keeps more distance from the human under the maximum distance preference.
    With a dynamic approaching human \textbf{(C+D)} that was not accounted for during training, the preference reflection decreases.
	}
	\label{fig:real_world}
\end{figure}
We evaluated our tuneable policy on a Kobuki TurtleBot 2 using ROS~\cite{quigley_ros_2009} and transferred the TurtleBot-trained policy to a Toyota Human Support Robot (HSR). 
The agent received ground truth human and goal positions, with the dynamic human localized via a Vive VR tracker. 
The HSR’s lidar, mounted in the front of its rotation center, may cause state space discrepancies for the policy.
Due to its $270^\circ$ coverage, compared to the TurtleBot’s $360^\circ$ lidar, the rear distance readings were filled with the maximum range of $4 \si{\meter}$.
The procedure ensures state consistency under the conservative assumption that rear obstacles are unlikely to impact navigation, as the robot can only move forward.
Another discrepancy arises in velocity command execution, both in sim-to-real transfer and between robots, due to differences in actuator dynamics and drive mechanisms, potentially affecting navigation performance.
We ran navigation tests on both robots for the preference vectors $\boldsymbol{\lambda} \in \Lambda$ with $N=5$ (see Sec.~\ref{sec:qualitative_analyis}).

The recorded TurtleBot trajectories are shown in Fig.~\ref{fig:real_world}.A and the HSR trajectories in Fig.~\ref{fig:real_world}.B. 
Both robots adapt their behavior according to preferences.
For the maximum distance preference (Fig.~\ref{fig:real_world}.A1), the TurtleBot shows oscillations, presumable due to slight over-steering, while the HSR drives closer to obstacles and exhibits a wider oscillatory motion near the goal (Fig.~\ref{fig:real_world}.B1).
These differences may result from lidar state mismatches (e.g., positional offset) or slower action execution due to inertia.
For maximum demonstration reflection, the TurtleBot’s trajectory aligns better with the demonstration than the HSR (Fig.~\ref{fig:real_world}.2).

Both robots avoid collisions with dynamically approaching humans (Fig.~\ref{fig:real_world}.C+D). 
As in the dynamic simulations experiments (Fig.~\ref{fig:qualitative_enhanced}), avoidance sharpens for the demonstration objective but fades as the human and robot pass each other. 
Under the distance preference, sharper inward steering and subsequent overshooting behind the human in simulation become more pronounced in the real world, compare (Fig.~\ref{fig:qualitative_enhanced}.C1+D1).
We attribute the sharper inward steering to the static training environment, which prevented the agent from learning in the presence of a moving human. Under static conditions, the agent typically maintains a fixed distance on the human’s side, forming a distance-angle mapping for avoidance. This mapping is disrupted by the dynamic human, causing the agent to turn inward as the human passes more quickly.
Efficiency-focused behavior transfers flawlessly.
Despite minor sim-to-real differences, all real-world trajectories remained collision-free, demonstrating robust sim-to-real generalization.
See the supplemental video for real-world experiments.\footnote{Supplemental video: \url{https://youtu.be/vS22B3HRdL4}}
In conclusion, the policy transfers smoothly to real robots, supporting research claim C4.

\section{Conclusion}
\label{sec:conclusion}
In summary, we introduced an innovative framework fusing multi-objective reinforcement learning~(MORL) with demonstration-based learning for adaptable, personalized robot navigation around a user with changing preferences. 
Our approach successfully modulates the conflicting objectives of demonstration data reflection, distance keeping, and navigational efficiency without retraining. 
To achieve this, we distill demonstration data into a reward model that shapes the agent's trajectories during navigation with variable strength. 
In various qualitative and quantitative experiments, we demonstrated the adaptability to varying preferences and scenarios.
Finally, we successfully deployed the learned agent on two real robots.

By accepting an externally controlled preference vector, the approach enables structured adaptation to changing user needs with a clear protocol for preference representation. 
Future research could focus on deriving such vectors from human feedback using a dedicated context-aware preference prediction agent.


\newpage
\bibliographystyle{IEEEtran}
\bibliography{bib_tuning_new,bib_context}

\end{document}